\RequirePackage{fix-cm}
\documentclass[smallextended]{svjour3}       % onecolumn (second format)
\smartqed  % flush right qed marks, e.g. at end of proof
\usepackage{graphicx}
\usepackage{algorithm,algpseudocode}

\usepackage{calc}
\usepackage{enumitem}
\usepackage{amsmath}
\begin{document}

\title{Topic modeling of behavioral modes using sensor data
\thanks{Invited Extended version of a paper \cite{resheffmatrix} presented at the international conference \textit{Data Science and Advanced Analytics}, Paris, France, 19-21 OCtober 2015 }
}



\author{Yehezkel S. Resheff         \and
        Shay Rotics  \and
        Ran Nathan \and
        Daphna Weinshall
}


\institute{Yehezkel S. Resheff \at
              Edmond and Lily Safra Center for Brain Sciences, The Hebrew University of Jerusalem, 91914, Israel \\              
              \email{yehezkel.resheff@mail.huji.ac.il}           %  \\
           \and Shay Rotics \at Movement Ecology Lab, Department of Ecology, Evolution and Behavior, The Hebrew University of Jerusalem
	       \and Ran Nathan \at Movement Ecology Lab, Department of Ecology, Evolution and Behavior, The Hebrew University of Jerusalem
           \and Daphna Weinshall \at School of Computer Science and Engineering, The Hebrew University of Jerusalem
}

\date{Received: date / Accepted: date}

\maketitle

\begin{abstract}
The field of Movement Ecology, like so many other fields, is experiencing a period of rapid growth in availability of data. As the volume rises, traditional methods are giving way to machine learning and data science, which are playing an increasingly large part it turning this data into science-driving insights. 
One rich and interesting source is the bio-logger. These small electronic wearable devices are attached to animals free to roam in their natural habitats, and report back readings from multiple sensors, including GPS and accelerometer bursts. A common use of accelerometer data is for supervised learning of behavioral modes. However, we need unsupervised analysis tools as well, in order to overcome the inherent difficulties of obtaining a labeled dataset, which in some cases is either infeasible or does not successfully encompass the full repertoire of behavioral modes of interest. 
Here we present a matrix factorization based topic-model method for accelerometer bursts, derived using a linear mixture property of patch features. Our method is validated via comparison to a labeled dataset, and is further compared to standard clustering algorithms.
\keywords{Behavioral Modes \and Topic Model \and Movement Ecology \and MS-BoP}
\end{abstract}

\section{Introduction}

Wearable devices with various sensors are becoming increasingly popular, with ongoing research into applications to health monitoring \cite{pantelopoulos2010survey} and context detection \cite{kern2003multi}. Many fields of animal behavior and conservation have also began to utilize similar devices in order to remotely monitor the whereabouts and behavior of their research subjects \cite{resheff2014accelerater}, and this has especially been the case in the field of Movement Ecology.         

The aim of Movement ecology is to unify research of movement of organisms and aid in the development of a general theory of whole-organism movement \cite{Nathan2008}. Recent technological advances in tracking tools and especially the appearance of cheap and small GPS devices \cite{Hebblewhite2010}, have driven the field into a period of rapid growth in knowledge and insight \cite{Holyoak2008}, and have led to the emergence of various methods of analyzing movement patterns \cite{Smouse2010}. 

Nevertheless, movement data, however accurate, is unlikely to suffice for inference on the links between behavioral, ecological, physiological, and evolutionary processes driving the movement of individuals, and link these subjects which have traditionally been researched separately in their respective fields. Thus, promoting movement ecology research and the desirable unification across species and movement phenomena requires the development of additional data sources: sensors and tools providing simultaneous information about the movement, energy expenditure and behavior of the focal organisms, together with the environmental conditions they encounter en route \cite{nathan2012using}.

One such tool, which has been introduced into the field of movement ecology, is the accelerometer-biologger (ACC). These sensors allow the determination of the acceleration of the tagged animal's body, and are used as a means of identifying moment-to-moment behavioral modes \cite{yoda1999precise}, and estimating energy expenditure \cite{wilson2006moving}.

ACC loggers typically record in 1-3 dimensions, either continuously or in short bouts in a constant window \cite{resheff2014accelerater}. Their output is used to infer behavior, most commonly through supervised machine learning techniques, and energy expenditure using the Overall Dynamic Body Acceleration (ODBA) or related metrics \cite{gleiss2011making,wilson2006moving}. When combined with GPS recordings, acceleration sensors add fine scale information on the variation in animal behavior, and energy expenditure in space and time (see \cite{Brown2013} for a recent review). 

ACC-based analysis has been used to compute many measures of interest in the field of Movement Ecology, including behavior-specific body posture, movement and activity budgets, measures of foraging effort, attempted food capture events, mortality detection, classifying behavioral modes and more \cite{Brown2013}. These measures have facilitated movement-related research for a wide range of topics in ecology and animal behavior \cite{Smouse2010,Brown2013,Takahashi2009,Spiegel2013} as well as other fields of research such as animal conservation and welfare \cite{Takahashi2009,Cooke2008} and biomechanics \cite{Hindle2010,Sellers2004}.

In recent years there has been considerable interest in the analysis of behavioral modes using ACC data and supervised learning techniques. The protocol for using ACC data for supervised learning of behavioral modes consists of several steps. First, a sensor calibration procedure is preformed in a controlled environment: before deployment, the response of each tag to $\pm1G$ acceleration on each axis is recorded, in order to fit the tag-specific linear transformation from the recorded values (mV) to the desired units of acceleration. Next, the calibrated tags are given a recording schedule and mounted on the focal animals, after these are captured. Finally, the data is retrieved using RF (radio) methods, Cellular transmission, or physically reacquiring the device. 

Once the data is retrieved, before supervised machine learning models can be used, a labeled dataset is collected through field observations. This time and labor intensive stage requires the researcher to observe the animal, either in its natural habitat or in captivity, and relate the actual behavioral modes to the time-stamp of the ACC recordings. Since some behavioral modes tend to be less common, or are performed predominantly at specific times, recording a sufficient number of such behavior-measurement samples may be tricky. Furthermore, for aquatic and nocturnal species, observations may not be feasible. In the final stage, models are trained using the labeled data, and the entire dataset is then labeled.  

Supervised machine learning methods have been applied to ACC data from many species, and for a diverse range of behavioral modes. However, there are several drawbacks to the supervised approach. Observations, even if perfectly accurate, may not be adequately representative of the behavioral pattern throughout the period of the research (which is desirably the lifetime of the animal), for several reasons: field work is inherently confined to a specific time and place; moreover, only some of the animals are observed, and the presence of the observer may in some cases have an impact on the behavior of the observed animals. Furthermore, the need for observations limits the scope of such research projects to observable species and to research labs with the necessary resources (in money, manpower, and knowledge) to carry out all the steps listed above.

In this paper we present a framework for unsupervised analysis of behavioral modes from ACC data. First we suggest a patch-codebook descriptor (MS-BoP) of ACC signals reminiscent of "bag of visual words" descriptors in Computer Vision (see \cite{csurka2004visual,zagoris2014distinction}). Next, we present a simple topic model for behavioral modes incorporating a linear mixture property of the MS-BoP features, and demonstrate how it can be used for unsupervised analysis of behavioral modes. 

The rest of the paper is organized as follows: The next section describes related work both in Movement Ecology and in matrix factorization for clustering and topic modeling. In section \ref{methodology-section} we introduce the features and model. Finally, in section \ref{results-section} we present the results of an analysis on a large real-world dataset and the comparison to other methods.

\section{Previous Work}

Previous work on behavioral mode analysis using ACC data focused predominantly on supervised learning, with an emphasis on constructing useful features and finding the right classifiers for a specific use, such as monitoring dairy cows \cite{diosdado2015classification}, or determining the flight type of soaring birds \cite{williams2015can}. 

While this line of work proved very successful, both in terms of classifier performance and of scientific discovery that it was able to drive, it still suffers from the inherent limitations of supervised learning, compounded by the very high cost of obtaining labeled data for behavioral observations of wild animals. It remains the case that for some animals (nocturnal or sea species for instance), obtaining a labeled dataset is currently infeasible. Thus, in order to use all available ACC data for behavioral mode analysis in the field of Movement Ecology, an unsupervised framework is essential.

To the best of our knowledge, there have been two attempts at such an approach. In \cite{Sakamoto2009}, K-means was applied to a representation of the ACC data, to achieve behavior-mode clusters. In \cite{garriga2015expectation,louzao2014coupling} a Gaussian Mixture Model (GMM) variant was used to cluster a low-dimensional representation of  ACC signals into a small number of useful behavioral modes. Our method goes one step further by allowing samples to be a mixture (more precisely, a convex combination) of behavioral modes, accounting for the observation that ACC samples do indeed tend to be mixed this way (Figure \ref{figure1}). 

Non-Negative Matrix Factorization (NNMF) has been studied extensively in the context of clustering \cite{wang2013nonnegative,li2006relationships} and topic modeling \cite{arora2012learning}. Connections have been shown to various popular clustering algorithms such as K-means and spectral clustering \cite{ding2005equivalence}. Our proposed method is essentially topic modeling with NNMF, based on theoretical justification that incorporates the nature of our signals and the features under consideration.

\section{Methodology}
\label{methodology-section}

\subsection{Feature generation}

In the field on Natural Language Processing (NLP), textual documents are commonly described as word-count histograms. These descriptors are generally known as \textit{bag-of-word} representations (BoW), since during their creation all the words in a document are (figuratively speaking) thrown into a bag, loosing all proximity information, then each word in a pre-defined dictionary is assigned the number of times it repeats in the bag. The final representation of the document is a vector of these counts.

The BoW representation was adopted in recent years into Computer Vision for the representation of images. Since images are not naturally divided into discrete elements (like words in a document), the first step is to transform the image into a series of word-analogues which can then be thrown into a bag. This discretization process is often achieved by clustering \textit{patches} of images, then assigning each patch the index of its cluster. The resulting feature vector for a given image is the histogram of the cluster associations of its patches. The cluster centroid are often referred to as the \textit{codebook}, and the method as Bag of Visual Words (BoVW).

Here, we adapt the BoVW method to be used with the ACC signal. We start by defining the notion of a patch of an ACC signal. 

\subsubsection{definition: patch in an ACC signal}
Let:
\[  s=[s_1,...,s_N] \] 
be an ACC signal of length $N$. The patch of length $l$ starting at index $i$ of $s$ is the sub-vector:
\[ [s_i,...,s_{i+l-1}] \]
thus, there are $N-l+1$ distinct patches in $s$.

\subsubsection{Codebook Generation}
As in the BoVW case, ACC signals and patches do not consist of discrete elements. In order to count and histogram types of patches, we must first construct a patch-codebook. 
We suggest the following construction: given a codebook size $k$ and a patch length $l$, for each ACC signal in the dataset, extract and pool all of the $l$-length-ed patches. Next, using K-means cluster the patches into $k$ clusters. The resulting $k$ centroids will be called the codebook.
The intuition behind using patches to describe an ACC signal, is that behavioral modes should be definable by the distribution of short-time-scale movements that they are comprised of. Since different behavioral modes occur at various characteristic timescales, we would like to repeat the process for more than one patch length, in order to efficiently capture all ACC patterns of relevance. 
Thus, we generate a separate codebook for several time-scales in the appropriate range, depending on the behavioral modes we are interested in (Alg. \ref{alg:cb-gen}).

\subsubsection{Feature Transformation}
Once we have constructed the codebook for all of the scales, we are ready to transform our ACC signals into the final Multi-Scale Bag of Patches (MS-BoP) descriptor. For each ACC record in the dataset, and for each scale, we extract all patches of the signal at that scale, and assign each one the index of the nearest centroid in the appropriate codebook. For each scale we then histogram the index values to produce a (typically sparse) vector of the length of the codebook. The final representation is the concatenation of histograms for the various scales (Alg. \ref{alg:MS-BoP-features}).

\begin{algorithm*}[tbh]
	\textbf{input:} 
	\begin{description}[leftmargin=!, labelwidth=\widthof{ ................. }]
		\item[$\{s_{i}\}_{i=1}^{p}$] the set of raw acceleration measurements
		\item[$l_{1},..,l_{m}$] list of scales to use 	
		\item[$k_{1},...,k_{m}$] list of corresponding sizes per codebook
	\end{description}
		  		
	\textbf{output:}
	\begin{description}[leftmargin=!, labelwidth=\widthof{ ................. }]
		\item[$CB_1,...,CB_l$] the generated codebooks. $CB_i[j]$ is the $j-th$ word in the $i-th$ codebook ($i=1,...,l; j=1,...,k_i$)
	\end{description}

	\begin{algorithmic}[1] 
				
		\For {scale := 1,...,l }
			\State {patches := list of all patches of scale \textit{$l_{scale}$} in $\{s_{i}\}_{i=1}^{p}$ }
			\State {$CB_i$ := K\_means(patches, $k_{scale}$).centroids}
		\EndFor

		\State {\Return {$CB_1,...,CB_l$}}
	\end{algorithmic}
	
	\protect\caption{Multi Scale Codebook Generation}
	\label{alg:cb-gen}
\end{algorithm*}

\begin{algorithm*}[tbh]
	\textbf{input:} 
	\begin{description}[leftmargin=!, labelwidth=\widthof{ ................. }]
		\item[$CB_1,...,CB_l$] The $l$ codebooks, output of Alg. \ref{alg:cb-gen}.
		\item[$l_{1},..,l_{m}$] list of the patch scales that were used in Alg. \ref{alg:cb-gen}.
		\item[$s$] an ACC signal to transform
	\end{description}
	
	\textbf{output:}
	\begin{description}[leftmargin=!, labelwidth=\widthof{ ................. }]
		\item[$f$] The MS-BoP representation of signal $s$		 
	\end{description}

	\begin{algorithmic}[1] 
		
		\For {scale := 1,...,l }
			\State {$f_{scale}$ := a zeros vector of the same length as $CB_{scale}$}
			
			\State {patches := list of all patches of scale \textit{$l_{scale}$} in $s$ }
			\ForAll {p in patches}
				\State {idx := index of the closest word to p in the codebook $CB_{scale}$ }
				\State {increment $f_{scale}[idx]$ by 1}
			\EndFor			 
		\EndFor
		
		\State {f := stack\_vectors($f_1,...,f_l$)}
		\State {\textbf{return:} f}
		
	\end{algorithmic}
	
	\protect\caption{MS-BoP feature transformation}
	\label{alg:MS-BoP-features}
\end{algorithm*}

\subsection{Mixture property of patch features}

In order to motivate the proposed model (next section), we present the mixture property of patch features. We assume that our signals have the property that a large enough part of a sample from a certain behavioral mode will have distribution of patches that is the same as the distribution in the entire sample. The meaning of this assumption is that each behavioral mode has a distribution of patches that characterizes it at each scale.

\begin{figure*}
	\includegraphics[width=0.95\textwidth]{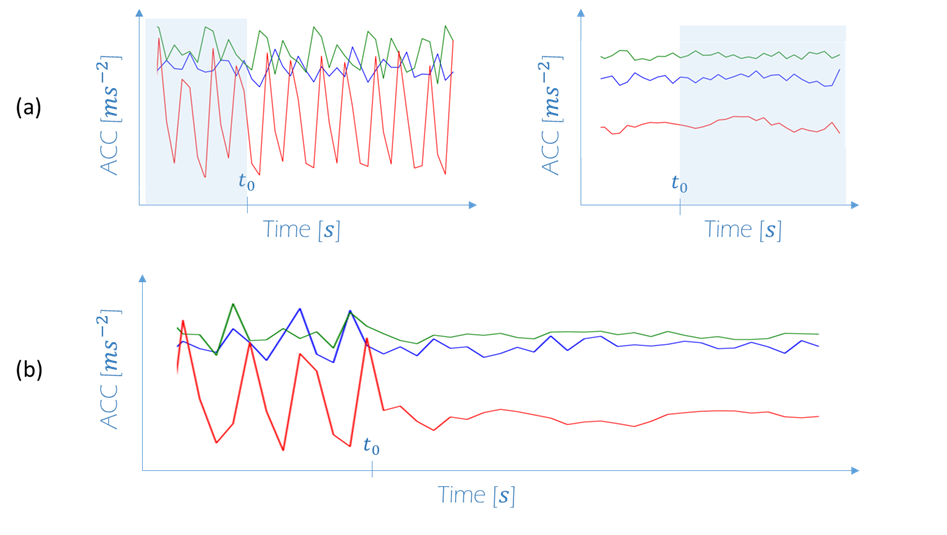}
	\caption{Pure and mixed triaxial ACC signals. Pure ACC signals (panel A) are measured during a single behavioral mode. However, in most cases a single measurement contains a mixture of more than one behavioral mode (Panel b), and may be viewed as a concatination of the beginning/end of two pure signals. The colors represent each of the three acceleration dimensions.}
	\label{figure1}
\end{figure*}

Intuitively, if a signal $s$ is constructed by taking the first half of a signal $s_{a}$ and the second half of an equal length signal $s_{b}$, then the distribution of patches in $s$ will be approximately an equal parts mixture of those in $s_{a}$ and in $s_{b}$. The reason for this is that a patch in $s$ is either (a) completely contained in $s_{a}$ and will then be distributed like patches in $s_{a}$ or, (b) completely in $s_{b}$, and will then be distributed like patches in $s_{b}$ or, (c) starts in $s_{a}$ and continues into $s_{b}$, in which case we know little about the patch distribution and consider it as noise. The key point is that the number of patches of type (c) is at most twice the length of the patch, and thus can be made small in relation to the total number of patches which is in the order of the length of the signal. More formally: 

Let $s$ be an ACC signal composed of a concatenation of $t_{1}$ consecutive samples during behavioral mode $a$ and $t_{2}$ consecutive samples during behavioral mode $b$ (see Figure \ref{figure1}). Denote $p_{mode}(v)$ the probability of a patch $v$ of length $l$ in behavioral $mode\in\{a,b\}$. Let $v$ be a patch drawn uniformly from $s$, then:

\begin{align*}
	p(v)&=Pr(A)p(v|A)+Pr(B)p(v|B)+Pr(C)p(v|C) \\
	&\geq Pr(A)p_{a}(v)+Pr(B)p_{b}(v) \\
	&=\frac{t_{1}-l}{t_{1}+t_{2}}p_{a}(v)+\frac{t_{2}-l}{t_{1}+t_{2}}p_{b}(v) \\
	&=\frac{t_{1}}{t_{1}+t_{2}}p_{a}(v)+\frac{t_{2}}{t_{1}+t_{2}}p_{b}(v)-\epsilon 
\end{align*}

\noindent where events $A,B,C$ denote the patch being all in $s_{1}$, all in $s_{2}$ and starting in $s{1}$ and ending in $s_{2}$ respectively, and:

\begin{equation*}
	\epsilon=\frac{l}{t_1+t_2}[p_{a}(v)+p_{b}(v)]
\end{equation*}

$\epsilon$ can be made arbitrarily small by making $t_{1}+t_{2}$ large and keeping $l$ constant, meaning that for patches small enough in relation to the length of the entire signal, the distribution of patches of the concatenated signal is a mixture (convex combination) of the distributions of the parts, with mixing coefficients proportional to the part lengths. We note that this result can easily be extended to a concatenation of any finite number of signals, as long as each one is sufficiently long in comparison to the patch width. 

Since behaviors of real-world animals may start and stop abruptly, and a recorded ACC signal is likely to be a concatenation of signals representing different behavioral modes (typically 1-3), the above property inspires a model that is able to capture such mixtures in an explicit fashion. Furthermore, the resulting mixture coefficients may provide some insight into the nature of the underlying behaviors and the relationships between them -- for example, which often appear alongside each other, and which are more temporally separated.

\subsection{The proposed model}

Let $k$ denote the number of behavioral modes under consideration, and $p$ the dimension of the representation of ACC observations. Following the mixture property presented in the previous section, we assume that every sample is a convex combination of the representation of a ``pure'' signal of the various behavioral modes. Further, we assume the existence of a matrix $F \in {R}^{pk}$, the \textit{factor matrix}, such that the $i-th$ column of $F$ is the representation of a pure signal of the $i-th$ behavioral mode, which we will call the factor associated with the $i-th$ behavioral mode. Let $s$ be an ACC sample, then:

\begin{equation}
	s=F\alpha+\epsilon\label{eq:one-sample-model}
\end{equation}

\noindent where $\epsilon \in R^p$ is some random vector. In other words, we say that the sample $s$ is a linear combination of the factors associated with each of the behavioral modes with some remainder term. For the full dataset, we then have:

\begin{equation}
	S=FA+\epsilon\label{eq:all-samples-model}
\end{equation}

\noindent where $F$ is the same matrix, $A's$ columns are the factor loadings for each of the samples denoted $\alpha$ in (\ref{eq:one-sample-model}), and $\epsilon \in R^{pN}$ is a random matrix. Since our features are non-negative histograms, and we would like the factor loadings to be non-negative, we constrain the matrices $F,A$ to have non-negative values. We solve for $F,A$ using a least squares criterion: 

\begin{equation}
	\begin{aligned}
		& \underset{F, A}{\text{argmin}}
		& & \Vert FA-S \Vert _{F}^2 \\
		& \text{subject to}
		& & F_{i,j},A_{i,j}\geq0\ \forall i,j
	\end{aligned}
	\label{eq:optimization-FA}
\end{equation}

\noindent This is by now a standard problem, which can be solved, for instance, using alternating non-negative least squares \cite{wang2013nonnegative}. The idea behind the algorithm  (Algorithm \ref{alg:alternating-nnls}) is that while the complete problem is not convex, and not easily solved, for a set $A$ it becomes a simple convex problem in $F$, and vice versa. This inspires the simple block-coordinate-descent algorithm which minimizes alternately w.r.t each of the matrices. Since this procedure generates a (weakly) monotonically decreasing series of values of the objective (\ref{eq:optimization-FA}), it is guaranteed to converge to a local minimum\footnote{The objective is bounded from below by $0$}. 

\begin{algorithm*}[tbh]
	\textbf{input:} 
	\begin{description}[leftmargin=!, labelwidth=\widthof{ ..... }]
		\item[$S$] the complete matrix $S \in R^{pN}$
		\item[$k$] factorization rank
	\end{description}
	
	\textbf{output:}
	\begin{description}[leftmargin=!, labelwidth=\widthof{ ..... }]
		\item[$F, A$] matrices $F \in R^{pk}$, $A \in R^{kN}$ 
	\end{description}
			
	\begin{algorithmic}[1] 
		
		\State{$F := $ random initialization}
		\State{$A := $ random initialization}
		\While{ not converged }
			\State{$ F := \underset{F}{\text{argmin}} \Vert FA-S \Vert _{F}^2 \ \ s.t. \  F_{i,j}\geq0\ \ \forall i,j $}
			\State{$ A := \underset{A}{\text{argmin}} \Vert FA-S \Vert _{F}^2 \ \ s.t. \ A_{i,j}\geq0\ \ \forall i,j $}
		\EndWhile
					
		\State {\Return {$F, A$}}
	\end{algorithmic}
	
	\protect\caption{Alternating Non-Negative Least Squares }
	\label{alg:alternating-nnls}
\end{algorithm*}

\subsection{Speed-up via sampling}

Since this method may potentially be applied to large datasets (containing at least hundreds of millions of records and many billions of patches), it is worth mentioning that all parameter-learning steps of the algorithm can be processed (identically to the original method) on a sample of the dataset. During codebook generation, records in the dataset and/or patches in each used record could be sampled to reduce the number of resulting patches we have to cluster. Next, fitting $F$ and $A$ on a sample of the records gives us the factor matrix, but not the factor loadings per record of the dataset. However, once we have $F$ the optimization problem  (\ref{eq:optimization-FA}) turns into:

\begin{equation}
	\begin{aligned}
		& \underset{A}{\text{argmin}}
		& & \Vert FA-S \Vert _{F}^2 \\
		& \text{subject to}
		& & A_{i,j}\geq0\ \forall i,j
	\end{aligned}
	\label{eq:optimization-A}
\end{equation}

\noindent a simple convex problem in which the factor loadings per record (columns of $A$) can be minimized independently for each record $s$ in the dataset, as follows:

\begin{equation}
	\begin{aligned}
		& \underset{\alpha}{\text{argmin}}
		& & \Vert F\alpha-s \Vert ^2 \\
		& \text{subject to}
		& & \alpha_{i}\geq0\ \forall i
	\end{aligned}
	\label{eq:optimization-alpha}
\end{equation}

\subsection{Extension to the multi-sensor case}

Thus far we have constructed a topic model applicable for data derived from a single (albeit possibly multi-dimensional) sensor. The multi-sensor (or sensor-integration) case is of particular interest in this case because many devices containing accelerometers also include other sensors such as gyroscopes and magnometers. Since each of these is recording at different frequencies, we can't simply consider them to be extra dimensions in the same time-series produced by the 3D accelerometer. 
The integrative framework we suggest assumes that the same behavioral modes are manifested in distinct patterns for each of the sensors. Thus, we will have separate factor matrices: 
\[ F^{1},...,F^{l} \] 
\noindent for the $l$ sensor types, and a single shared factor loading matrix $A$.
Denoting the features matrices of the MS-BoP features for each of the $l$ sensor types:
\[ S^{1},...,S^{l} \]
\noindent we now look for matrices:  
\[ A,F^{1},...,F^{l} \]
\noindent such that: 
\[ \forall i: S^{i} \approx F^{i}A \]
\noindent which we encode in the following optimization problem:

\begin{equation}
	\begin{aligned}
		& \underset{F^{1},...,F^{l}, A}{\text{argmin}}
		& & \frac{1}{l}\sum_{i=1}^{l}\Vert F^iA-S^i \Vert _{F}^2 \\
		& \text{subject to}
		& & F^k_{i,j},A_{i,j}\geq0\ \forall i,j, k	
	\end{aligned}
	\label{eq:optimization-FA-multi}
\end{equation}

\noindent This problem is solvable using the same type of method. Specifically,  we will now show that this new problem can be re-written in the same form as (\ref{eq:optimization-FA}), with both the sample and factor matrices stacked. Denote:

\[ 
F = \begin{bmatrix}
[F^1] \\
\vdots \\
[F^l] 
\end{bmatrix}
\] 

\noindent and:

\[ 
S = \begin{bmatrix}
[S^1] \\
\vdots \\
[S^l] 
\end{bmatrix}
\] 

\noindent then (\ref{eq:optimization-FA-multi}) becomes::

\[ 
\begin{aligned}
	& \underset{F, A}{\text{argmin}}
		& & \Vert FA-S \Vert _{F}^2 \\
	& \text{subject to}
		& & F_{i,j},A_{i,j}\geq0\ \forall i,j
\end{aligned}
\] 

\noindent since the $\frac{1}{l}$ scaling factor makes no difference to the $argmin$. 
In summary, the multi-sensor case where a separate factor matrix is allocated to each sensor, with a joint factor-loading matrix, is identical to the single-sensor case when the MS-BoP features for each sensor are stacked vertically.

\subsection{Extension to the supervised and semi-supervised cases}

Supposing observation (or any other mechanism) allowed us to obtain "pure" ACC signals for some (or all) of the behavioral modes. Using the mean MS-BoP representation of the signals in each of these modes for the corresponding column of $F$, we are left with a convex problem similar to (\ref{eq:optimization-FA}), where the optimization is over the remaining elements of $F$ only.

In the extreme case, when we have labeled samples for a pure ACC signal for all the behavioral modes under consideration, and thus all of $F$ is predetermined, the resulting problem is equivalent to (\ref{eq:optimization-A}). Namely, we are left with the task of obtaining the factor loadings for the remaining (unlabeled) data. 

\subsection{Limitations}

Consider a solution, matrices $F,A$ that minimize objective (\ref{eq:optimization-FA}), so that:
 \[ S \approx FA \]
 Clearly, for any Orthogonal matrix $Q$ (of the appropriate dimensions):
 \[ FA = FQQ^TA = (FQ)(A^TQ)^T\]
 \noindent thus, the solution:
\begin{align}
&  F' = FQ  \notag\\
&  A' = (A^TQ)^T \notag
\end{align}
 \noindent is also a minimizer of objective (\ref{eq:optimization-FA}), iff the matrices $F', A'$ obey the constraints:
\begin{equation}
	 F'_{i,j}, A'_{i,j} \geq 0  \ \forall i,j
	 \label{F,A-all-nn} 
\end{equation}

While this clearly holds if $Q$ is a permutation matrix, there are (always) orthogonal matrices $Q$ which contain negative elements for which the constraints in (\ref{F,A-all-nn}) hold. From the construction of $F'$ and $A'$, we can interpret them as an entanglement of our factors and loadings (technically, what we find is the span of the correct factors, but not the factors themselves). We note that while this property limits the ability to recover factors that generate the data, in practice the factors themselves are useful for analysis of behavioral topics, as demonstrated in the section below.

We leave to future research the issue of the disentanglement, which should be achieved via regularization with respect to $A$ in the original optimization problem (\ref{eq:optimization-FA}).

\section{Results}
\label{results-section}

In this section we present experiments designed to compare our method to alternatives, and derive insights about the data. Results are then discussed in the next section.

Data for these experiments consists of $3D$ acceleration measurements from bio-loggers which were recorded during $2012$. Each measurement consists of $4$ seconds at $10Hz$ per axis, giving a total of $120$ values. 

A ground truth partitioning of the data was obtained using standard machine learning techniques (see \cite{resheff2014accelerater,nathan2012using} for more details regarding the methodology), based on $3815$ field observations each of which was assigned one of $5$ distinct behavioral modes (Walking, Standing, Sitting, Flapping, Gliding). Experiments were conducted using stratified sampling of $100,000$ measurements ($20,000$ per behavioral mode). 

Matrix factorization was preformed using the scikit-learn \cite{scikit-learn} python software library (see \cite{lin2007projected} for method details). In all experiments the results were stable across repetitions, leading to essentially zero standard deviation, and therefore the reported results correspond to single repetitions. 

The purpose of these experiments is to assess to what extent the soft-partitioning via our topic model method relates to the hard, ground truth partitions. Our method is compared to the following: 

\begin{description}
	\item[\textbf{Random partitioning:}] each sample is assigned a value drawn uniformly
	from the set of possible partitions $\{1,2,..,k\}$
	
	\item[\textbf{Uniform partition:}] each sample is assigned the same distribution
	of $\frac{1}{k}$ per partition, over the $k$ partitions. 
	
	\item[\textbf{Kmeans:}] the sample are partitioned using Kmeans.
	
	\item[\textbf{Gaussian Mixture Models (GMM):}] GMM is used to assign samples $k$ partition coefficients. 
	 
\end{description}

\noindent where (a) and (b) are used as controls, (c) and (d) are used as representative hard and soft clustering methods, respectively.

\begin{figure*}
	\includegraphics[width=0.95\textwidth]{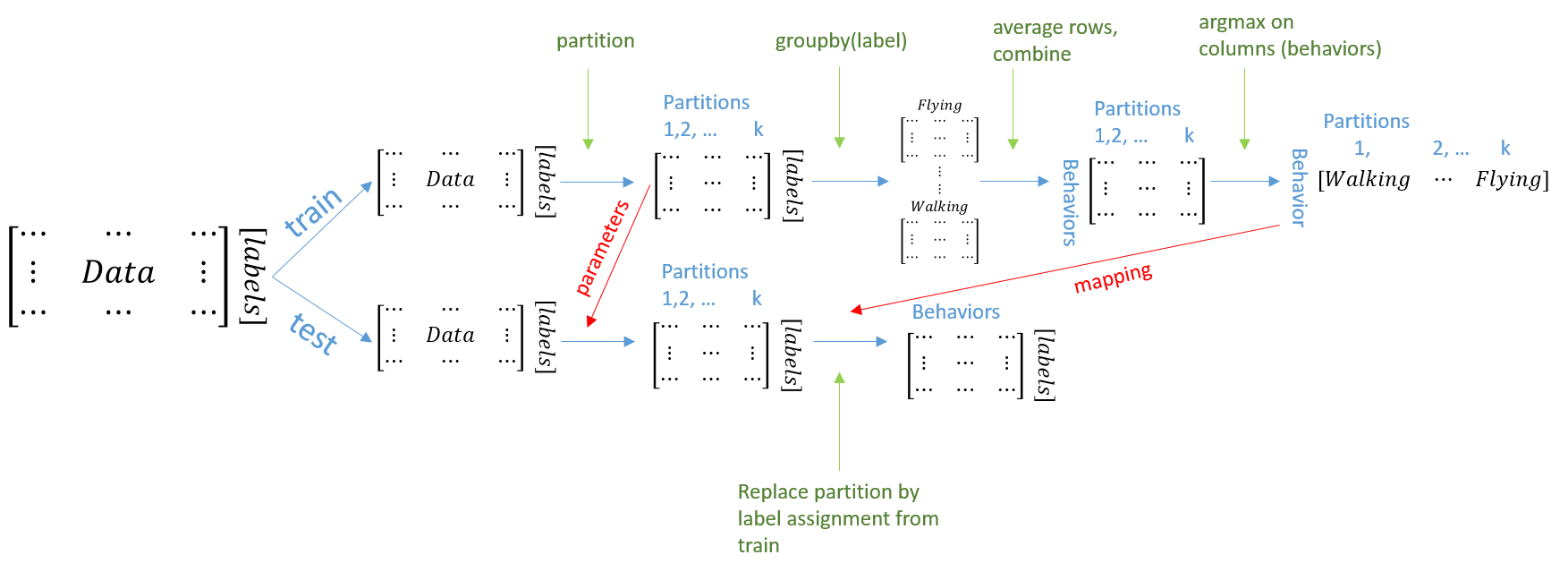}
	\caption{Schematic flow of partition evaluation}
	\label{figure2}
\end{figure*}

The data is then divided randomly into two equal parts designated train and test. Using the training-set we learn the partitioning of the data for each of the methods (random, uniform, Kmeans, GMM, and NNMF). Next, for each method separately, we assign each of the partitions one of the semantic labels (Flapping, Gliding, Walking, Standing, Sitting). In order to do this we group the data in the training-set according to the semantic label it received (the supervised annotation), and compute the average loading for each label in the partition. The final assignment for the partition is the label with the highest mean loading in it (see schematic in Fig. \ref{figure2}). 

\begin{figure}[htbp]
	\begin{centering}
		\textsf{\includegraphics[width=1\columnwidth]{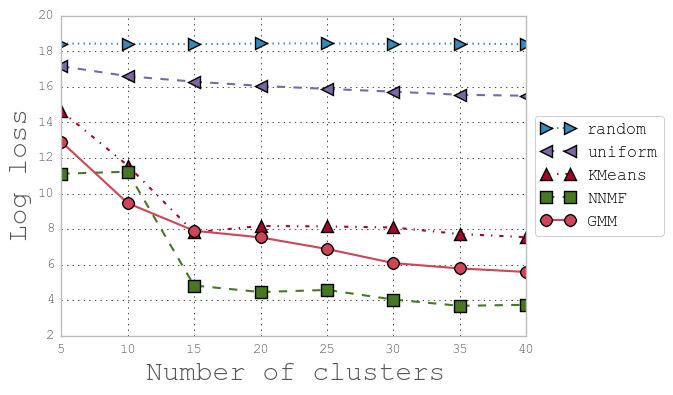}}
		\par\end{centering}
	
	\protect\caption{Log loss of soft-assignment to each of the ground-truth classes using
		each of the methods under consideration. (NNMF: non-negative matrix
		factorization, GMM: Gaussian mixture model)}
	\label{log-loss-figure}
\end{figure}

\begin{figure}[htbp]
	\begin{centering}
		\textsf{\includegraphics[width=1\columnwidth]{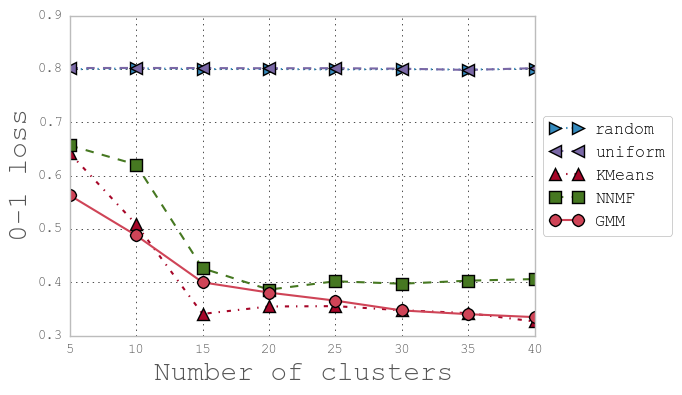}}
		\par\end{centering}
	
	\protect\caption{0-1 loss of hard-assignment to each of the ground-truth classes using
		each of the methods under consideration. For the soft-assignment partitioning
		methods, hard-assignment is achieved using argmax. (NNMF: non-negative
		matrix factorization, GMM: Gaussian mixture model)}
	\label{0-1-loss-figure}
\end{figure}

The evaluation stage is preformed on the test-set only. Resemblance to the ground-truth is measured using log-loss (Figure \ref{log-loss-figure}) and $0-1$ loss (Figure \ref{0-1-loss-figure}), after partition values are converted to soft label assignments using the mapping derived from the training set (see schematic in Fig. \ref{figure2}). 
For an assignment $l_1,...,l_5$ for the $5$ behavior labels, where the ground-truth label is $i$, we use the $0-1$ loss:

\begin{equation}
	l_{0-1} = \begin{cases}
		0 & i=argmax \{l_1,...,l_5\} \\
		1 & otherwise
	\end{cases}
\end{equation}

\noindent and the log-loss:

\begin{equation}
	l_{log} = -log(l_i)
\end{equation}

Table 1 shows the average distribution of supervised (ground-truth) behavioral modes for partitions assigned each of the labels, in the form of a confusion matrix. Partitions were obtained using non-negative matrix-factorization (NNMF) with $k=30$, and associations between partitions and labels as described above. Data is presented after row normalization to facilitate between-row comparison. 

\begin{table*}[tp]
	\protect\caption{Mean label association per ground-truth behavioral mode. NNMF with
		30 factors. Normalized rows.}	
	
	\centering{}%
	\begin{tabular}{|l|l|l|l|l|l|}
		\hline 
		Ground truth / Assignment & Flapping & Gliding & Walking & Standing & Sitting\tabularnewline
		\hline 
		\hline 
		Flapping & \textbf{51.25\%} & 13.66\% & 13.37\% & 4.33\% & 17.39\%\tabularnewline
		\hline 
		Gliding & 0.75\% & \textbf{49.98\%} & 8.49\% & 3.95\% & 36.84\%\tabularnewline
		\hline 
		Walking & 2.41\% & 19.71\% & \textbf{43.92\%} & 20.56\% & 13.41\%\tabularnewline
		\hline 
		Standing & 0.86\% & 13.30\% & 1.04\% & \textbf{74.93\%} & 9.88\%\tabularnewline
		\hline 
		Sitting & 0.01\% & 30.88\% & 0.15\% & 10.46\% & \textbf{58.50\%}\tabularnewline
		\hline 
	\end{tabular}
\end{table*}

\section{Discussion}

As expected, both $0-1$ and log-loss error plots are monotonically decreasing in the number of clusters (we use the term clusters here for cluster/partition/topic depending on the method under consideration). The most striking result is that while the matrix-factorization topic model method preforms well compared to the other methods with respect to the log-loss metric (Figure \ref{log-loss-figure}), it is not quite as good with respect to the 0-1 loss (Figure \ref{0-1-loss-figure}). 

In order to better understand this phenomena, we take a closer look at the data. Consider an observation where the animal takes a single step during the $4$-second acceleration measurement window, and stands still for the rest of it. In order not to dramatically underestimate the amount of walking, an observer will label this sample as Walking (In fact, most samples are probably mixtures). 

From the mixture property of the MS-BoP features (see Methodology section), when using the matrix factorization topic model approach we would expect to get a Walking factor proportional to the time spent doing so in the measurement windows. Thus, for a sample with some walking (say, less than 50\%) we get a miss in the 0-1 loss metric, but a better score in the log-loss which is more sensitive to assignment of low probabilities to the correct class. 

Table 1 sheds more light on the aforementioned result by showing average assignment of factors for each of the ground-truth classes, in the form a confusion matrix. Flapping samples indeed received the highest weight, on average, on Flapping factors (51.25\%), but the Gliding and Walking factors get over 13\% each. This may be due to the fact that Storks indeed glide between wing flaps, and may have walked prior to taking off during the observations which are inherently biased to behavior close to the ground (where the observer is). Conversely, none of the other behavioral modes include a significant amount of Flapping factors. 

This result may also point to the tendency (or strategy) of field observers to assign the more active behavior to mixed samples (In which case a sample where the bird flaps for a part of the duration of the measurement would be assigned to Flapping, in the same sense that a step or two would qualify an otherwise stationary sample as Walking). 

We note that the Sitting factors received factor weights higher than expected in all other behavioral modes. It might be interesting to try and overcome this sort of systematic error using a column normalization. We defer this to future research.

\section{Conclusions}

In this paper we describe a matrix factorization based topic model approach to behavioral mode analysis from accelerometer data and demonstrate its qualities using a large Movement Ecology dataset. While clustering and topic modeling with matrix factorization is by no means a new idea, the novelty here is in the integration with patch features (MS-BoP) that theoretically motivate the method in the context of time-series sensor readings for behavioral mode analysis. 

The main contribution of this paper is in presenting a framework that will allow for a widespread use of behavioral mode analysis in Movement Ecology, and related fields where determining movement patterns from remote sensor readings is necessary. Further, we introduce the MS-BoP features, which may be applicable for many continuous sensor readings, and show that a linear mixture model is justified when using such features.

\section*{Acknowledgment}

This work was supported in part by a grant from the Israel Science Foundation (ISF) to Prof. Daphna Weinshall.

\bibliographystyle{abbrv}
\bibliography{lib}

\begin{thebibliography}{10}

\bibitem{arora2012learning}
S.~Arora, R.~Ge, and A.~Moitra.
\newblock Learning topic models--going beyond svd.
\newblock In {\em Foundations of Computer Science (FOCS), 2012 IEEE 53rd Annual
  Symposium on}, pages 1--10. IEEE, 2012.

\bibitem{Brown2013}
D.~D. Brown, R.~Kays, M.~Wikelski, R.~Wilson, and a.~Klimley.
\newblock {Observing the unwatchable through acceleration logging of animal
  behavior}.
\newblock {\em Animal Biotelemetry}, 1(1):20, 2013.

\bibitem{Cooke2008}
S.~Cooke.
\newblock {Biotelemetry and biologging in endangered species research and
  animal conservation: relevance to regional, national, and IUCN Red List
  threat assessments}.
\newblock {\em Endangered Species Research}, 4(January):165--185, Jan. 2008.

\bibitem{csurka2004visual}
G.~Csurka, C.~Dance, L.~Fan, J.~Willamowski, and C.~Bray.
\newblock Visual categorization with bags of keypoints.
\newblock {\em Workshop on statistical learning in computer vision, ECCV},
  1(1-22):1--2, 2004.

\bibitem{ding2005equivalence}
C.~H. Ding, X.~He, and H.~D. Simon.
\newblock On the equivalence of nonnegative matrix factorization and spectral
  clustering.
\newblock In {\em SDM}, volume~5, pages 606--610. SIAM, 2005.

\bibitem{diosdado2015classification}
J.~A.~V. Diosdado, Z.~E. Barker, H.~R. Hodges, J.~R. Amory, D.~P. Croft, N.~J.
  Bell, and E.~A. Codling.
\newblock Classification of behaviour in housed dairy cows using an
  accelerometer-based activity monitoring system.
\newblock {\em Animal Biotelemetry}, 3(1):15, 2015.

\bibitem{garriga2015expectation}
J.~Garriga, J.~R. Palmer, A.~Oltra, and F.~Bartumeus.
\newblock Expectation-maximization binary clustering for behavioural
  annotation.
\newblock {\em arXiv preprint arXiv:1503.04059}, 2015.

\bibitem{gleiss2011making}
A.~C. Gleiss, R.~P. Wilson, and E.~L.~C. Shepard.
\newblock {Making overall dynamic body acceleration work: on the theory of
  acceleration as a proxy for energy expenditure}.
\newblock {\em Methods in Ecology and Evolution}, 2(1):23--33, 2011.

\bibitem{Hebblewhite2010}
M.~Hebblewhite and D.~T. Haydon.
\newblock {Distinguishing technology from biology: a critical review of the use
  of GPS telemetry data in ecology.}
\newblock {\em Philosophical transactions of the Royal Society of London.
  Series B, Biological sciences}, 365(1550):2303--12, July 2010.

\bibitem{Hindle2010}
A.~Hindle, D.~Rosen, and A.~Trites.
\newblock {Swimming depth and ocean currents affect transit costs in Steller
  sea lions Eumetopias jubatus}.
\newblock {\em Aquatic Biology}, 10(2):139--148, Aug. 2010.

\bibitem{Holyoak2008}
M.~Holyoak, R.~Casagrandi, R.~Nathan, E.~Revilla, and O.~Spiegel.
\newblock {Trends and missing parts in the study of movement ecology.}
\newblock {\em Proceedings of the National Academy of Sciences of the United
  States of America}, 105(49):19060--5, Dec. 2008.

\bibitem{kern2003multi}
N.~Kern, B.~Schiele, and A.~Schmidt.
\newblock Multi-sensor activity context detection for wearable computing.
\newblock In {\em Ambient Intelligence}, pages 220--232. Springer, 2003.

\bibitem{li2006relationships}
T.~Li and C.~Ding.
\newblock The relationships among various nonnegative matrix factorization
  methods for clustering.
\newblock In {\em Data Mining, 2006. ICDM'06. Sixth International Conference
  on}, pages 362--371. IEEE, 2006.

\bibitem{lin2007projected}
C.-b. Lin.
\newblock Projected gradient methods for nonnegative matrix factorization.
\newblock {\em Neural computation}, 19(10):2756--2779, 2007.

\bibitem{louzao2014coupling}
M.~Louzao, T.~Weigand, F.~Bartumeus, and H.~Weimerskirch.
\newblock Coupling instantaneous energy-budget models and behavioural mode
  analysis to estimate optimal foraging strategy: an example with wandering
  albatrosses.
\newblock {\em Mov Ecol}, 2(8), 2014.

\bibitem{Nathan2008}
R.~Nathan and W.~Getz.
\newblock {A movement ecology paradigm for unifying organismal movement
  research}.
\newblock {\em Proceedings of the National Academy of Sciences of the United
  States of America}, 105(49):19052--19059, 2008.

\bibitem{nathan2012using}
R.~Nathan, O.~Spiegel, S.~Fortmann-Roe, R.~Harel, M.~Wikelski, and W.~M. Getz.
\newblock {Using tri-axial acceleration data to identify behavioral modes of
  free-ranging animals: general concepts and tools illustrated for griffon
  vultures}.
\newblock {\em The Journal of experimental biology}, 215(6):986--996, 2012.

\bibitem{pantelopoulos2010survey}
A.~Pantelopoulos and N.~G. Bourbakis.
\newblock A survey on wearable sensor-based systems for health monitoring and
  prognosis.
\newblock {\em Systems, Man, and Cybernetics, Part C: Applications and Reviews,
  IEEE Transactions on}, 40(1):1--12, 2010.

\bibitem{scikit-learn}
F.~Pedregosa, G.~Varoquaux, A.~Gramfort, V.~Michel, B.~Thirion, O.~Grisel,
  M.~Blondel, P.~Prettenhofer, R.~Weiss, V.~Dubourg, J.~Vanderplas, A.~Passos,
  D.~Cournapeau, M.~Brucher, M.~Perrot, and E.~Duchesnay.
\newblock Scikit-learn: Machine learning in {P}ython.
\newblock {\em Journal of Machine Learning Research}, 12:2825--2830, 2011.

\bibitem{resheff2014accelerater}
Y.~S. Resheff, S.~Rotics, R.~Harel, O.~Spiegel, and R.~Nathan.
\newblock {AcceleRater: a web application for supervised learning of behavioral
  modes from acceleration measurements}.
\newblock {\em Movement Ecology}, 2(1):25, 2014.

\bibitem{resheffmatrix}
Y.~S. Resheff, S.~Rotics, R.~Nathan, and D.~Weinshall.
\newblock Matrix factorization approach to behavioral mode analysis from
  acceleration data.
\newblock In {\em Data Science and Advanced Analytics (DSAA), 2015
  International Conference on}. IEEE, 2015.

\bibitem{Sakamoto2009}
K.~Q. Sakamoto, K.~Sato, M.~Ishizuka, Y.~Watanuki, A.~Takahashi, F.~Daunt, and
  S.~Wanless.
\newblock {Can ethograms be automatically generated using body acceleration
  data from free-ranging birds?}
\newblock {\em PloS one}, 4(4):e5379, Jan. 2009.

\bibitem{Sellers2004}
W.~I. Sellers and R.~H. Crompton.
\newblock {Automatic monitoring of primate locomotor behaviour using
  accelerometers.}
\newblock {\em Folia primatologica; international journal of primatology},
  75(4):279--93, 2004.

\bibitem{Smouse2010}
P.~E. Smouse, S.~Focardi, P.~R. Moorcroft, J.~G. Kie, J.~D. Forester, and J.~M.
  Morales.
\newblock {Stochastic modelling of animal movement.}
\newblock {\em Philosophical transactions of the Royal Society of London.
  Series B, Biological sciences}, 365(1550):2201--11, July 2010.

\bibitem{Spiegel2013}
O.~Spiegel, R.~Harel, W.~M. Getz, and R.~Nathan.
\newblock {Mixed strategies of griffon vultures (Gyps fulvus) response to food
  deprivation lead to a hump-shaped movement pattern}.
\newblock {\em Movement Ecology}, 1(1):5, 2013.

\bibitem{Takahashi2009}
M.~Takahashi, J.~R. Tobey, C.~B. Pisacane, and C.~H. Andrus.
\newblock {Evaluating the utility of an accelerometer and urinary hormone
  analysis as indicators of estrus in a Zoo-housed koala (Phascolarctos
  cinereus).}
\newblock {\em Zoo biology}, 28(1):59--68, 2009.

\bibitem{wang2013nonnegative}
Y.-X. Wang and Y.-J. Zhang.
\newblock Nonnegative matrix factorization: A comprehensive review.
\newblock {\em Knowledge and Data Engineering, IEEE Transactions on},
  25(6):1336--1353, 2013.

\bibitem{williams2015can}
H.~Williams, E.~Shepard, O.~Duriez, and S.~Lambertucci.
\newblock Can accelerometry be used to distinguish between flight types in
  soaring birds?
\newblock {\em Animal Biotelemetry}, 3(1):1--11, 2015.

\bibitem{wilson2006moving}
R.~P. Wilson, C.~R. White, F.~Quintana, L.~G. Halsey, N.~Liebsch, G.~R. Martin,
  and P.~J. Butler.
\newblock {Moving towards acceleration for estimates of activity-specific
  metabolic rate in free-living animals: the case of the cormorant}.
\newblock {\em Journal of Animal Ecology}, 75(5):1081--1090, 2006.

\bibitem{yoda1999precise}
K.~Yoda, K.~Sato, Y.~Niizuma, M.~Kurita, C.~Bost, Y.~{Le Maho}, and Y.~Naito.
\newblock {Precise monitoring of porpoising behaviour of Ad{\'e}lie penguins
  determined using acceleration data loggers}.
\newblock {\em Journal of Experimental Biology}, 202(22):3121--3126, 1999.

\bibitem{zagoris2014distinction}
K.~Zagoris, I.~Pratikakis, A.~Antonacopoulos, B.~Gatos, and N.~Papamarkos.
\newblock Distinction between handwritten and machine-printed text based on the
  bag of visual words model.
\newblock {\em Pattern Recognition}, 47(3):1051--1062, 2014.

\end{thebibliography}

\end{document}